\documentclass[lettersize,journal]{IEEEtran}
\usepackage{amsmath,amsfonts}
\usepackage{algorithmic}
\usepackage{algorithm}
\usepackage{array}
\usepackage[caption=false,font=normalsize,labelfont=sf,textfont=sf]{subfig}
\usepackage{textcomp}
\usepackage{stfloats}
\usepackage{url}
\usepackage{verbatim}
\usepackage{graphicx}
\usepackage{cite}
\usepackage{amssymb}
\usepackage{booktabs}
 \usepackage[switch]{lineno}

\begin{document}
	

\title{Multilateral Cascading Network for Semantic Segmentation \\of Large-Scale Outdoor Point Clouds }

\author{Haoran Gong, Haodong Wang, and Di Wang
\thanks{This work is supported by the National Natural Science Foundation of China (No. 42101330), the National Key Research and Development Program of China (No. 2021YFF0704600), the Key Research and Development Program of Shaanxi Province (No. 2023-YBSF-452). \textit{(Corresponding author: Di Wang.)}}
\thanks{Haoran Gong, and Di Wang are with the School of Software Engineering, Xi’an Jiaotong University, Xi'an 710049, China (email: gonghr@stu.xjtu.edu.cn; diwang@mail.xjtu.edu.cn).}
\thanks{Haodong Wang is with the School of Electronic Engineering, Xidian University, Xi'an 710071, China (email: 21021210997@stu.xidian.edu.cn).}

}



\maketitle

\begin{abstract}
Semantic segmentation of large-scale outdoor point clouds is of significant importance in environment perception and scene understanding. However, this task continues to present a significant research challenge, due to the inherent complexity of outdoor objects and their diverse distributions in real-world environments. In this study, we propose the Multilateral Cascading Network (MCNet) designed to address this challenge. The model comprises two key components: a Multilateral Cascading Attention Enhancement (MCAE) module, which facilitates the learning of complex local features through multilateral cascading operations; and a Point Cross Stage Partial (P-CSP) module, which fuses global and local features, thereby optimizing the integration of valuable feature information across multiple scales. Our proposed method demonstrates superior performance relative to state-of-the-art approaches across two widely recognized benchmark datasets: Toronto3D and SensatUrban. Especially on the city-scale SensatUrban dataset, our results surpassed the current best result by 2.1\% in overall mIoU and yielded an improvement of 15.9\% on average for small-sample object categories comprising less than 2\% of the total samples, in comparison to the baseline method.
\end{abstract}


\section{Introduction}
\IEEEPARstart{S}emantic segmentation is a critical task of leveraging point cloud data, which involves classifying each point into meaningful categories. This process enables a deeper understanding of spatial environments, facilitating improved object recognition \cite{qi2021offboard}, scene interpretation \cite{moyano2021semantic}, and navigation \cite{el2021indoor}. 

The rapid advancement of deep learning has led to significant advances in point cloud semantic segmentation in the field of remote sensing. The groundbreaking development of PointNet \cite{pointnet} enabled the direct processing of 3D point clouds. Since its introduction, various advanced neural networks based on point clouds have been proposed \cite{pointnet++,dgcnn,pointnext}. These methods have greatly facilitated the development of point cloud processing techniques in the context of deep learning, especially semantic segmentation. Early works mainly focused on small-scale datasets and scenes. 

As research progresses, attention has shifted from small-scale to large-scale outdoor point cloud semantic segmentation, where challenges arise due to the complex diversity of outdoor objects in size, shape, and distribution. This makes feature learning difficult, and class imbalance further complicates model training.

A variety of methods have been developed and evaluated to address these challenges. For example, SPG \cite{spg} addresses the issue of managing large data sets by converting extensive point clouds into hypergraphs prior to training the model. Nevertheless, this hypergraph preprocessing introduces a considerable computational burden. To address this challenge, RandLA-Net \cite{randla} was introduced as a lightweight network that employs random downsampling and local feature aggregation blocks, thereby enhancing segmentation performance while significantly reducing training costs. Similarly, KPConv \cite{kpconv} is a kernel-based convolutional network designed for point clouds, effectively extending convolutional operations to 3D data. The network demonstrates particular efficacy in learning specific features of 3D structural objects, as evidenced by its success in segmenting objects with broad applicability. More recent work such as NeiEA-Net \cite{neiea} focuses on maximizing high-dimensional feature spaces and optimizing local neighborhoods in 3D Euclidean space, enabling better capture of local details. LEARD-Net \cite{leard} employs multi-hop connections and color coding to learn contextual information. LACV-Net \cite{zeng2024large} introduces a Local Adaptive Feature Augmentation (LAFA) module that adaptively learns similarity weights among local neighbors, enhancing local information and minimizing ambiguity. Transformer-based method \cite{lai2022stratified,10273676} benefits from capturing long-range dependencies between points, but may suffer from computational cost as point cloud size increases.

\begin{figure*}
	\begin{center}
		\includegraphics[width=1.0\textwidth]{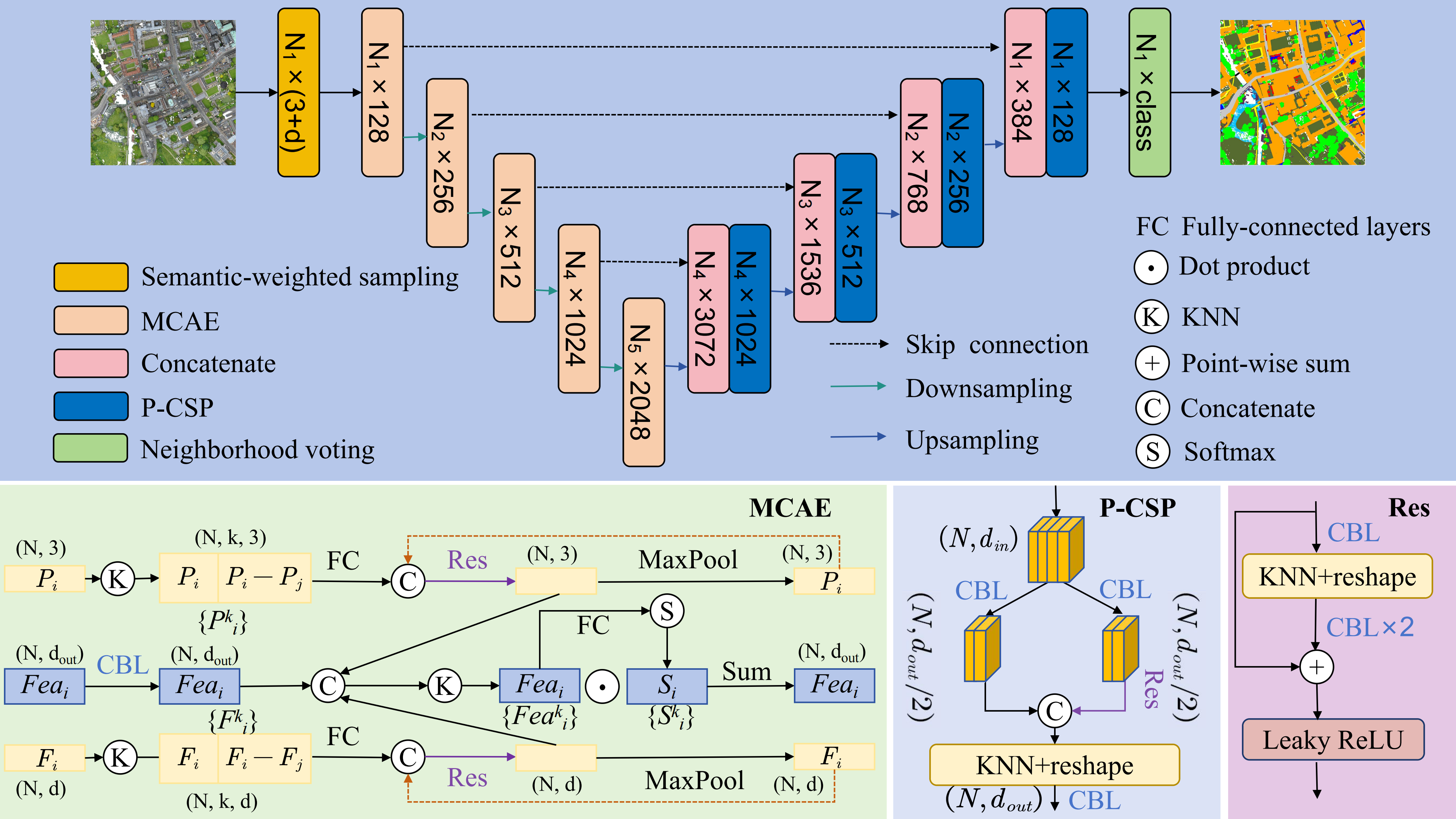}
	\end{center}
	\caption{ \textbf{Overall Architecture}. "CBL" means features going through 1$\times$1 convolution, batch normalization and Leaky ReLU.}
	\label{fig:architecture}
\end{figure*}

Despite the rapid development of this field, the majority of current methods treat the coordinate and feature information of the point cloud in a uniform manner by stacking them along the channel dimension before feeding them into the network. While this approach is straightforward, it fails to fully exploit the distinct characteristics of coordinate and feature information. In contrast, our method addresses this limitation by independently extracting features from coordinates and feature information first, followed by their integration at a later stage. This strategy of separate extraction and subsequent fusion effectively captures the complementary properties of both types of information, significantly enhancing the overall feature extraction capability. 

In summary, our main contributions are:

\begin{enumerate}
	
	\item[$\bullet$] A Multilateral Cascading Attention Enhancement (MCAE) module is introduced at the encoding layer, employing a flexible cascading approach.
	
	\item[$\bullet$] A Point Cross Stage Partial (P-CSP) module is designed at the decoding layer as a novel feature aggregation strategy to optimize the integration of valuable feature information across all scales.

	\item[$\bullet$] We present a new network, termed MCNet, achieving state-of-the-art performance in semantic segmentation across two benchmarks.
	
\end{enumerate}
 
\section{Methodology}

\subsection{Overall Network Architecture}

\subsubsection{Semantic-weighted Sampling Module}
To alleviate sample imbalance, a weighting mechanism is applied to the training data, based on its semantic information. A probability parameter is introduced for each point based on the distance ratio between the point and its corresponding central point. Then the weight of each label type according to its proportion within the overall sample is calculated and integrated with the parameter. 

\subsubsection{Encoding Layer}
The encoding layer is composed of MCAE modules. The MCAE is designed to extract local feature in a multilateral cascading manner, which treats the coordinate and feature information of point clouds independently and integrates them at a later stage. Moreover, feature maps at each level are delivered to the decoding layer at the same level through skip connection.

\subsubsection{Decoding Layer}
We design the P-CSP module at the decoding layer to fuse sampled features at the same scale. First, features from last level are upsampled through transpose convolution. Next, features are concatenated with those from the encoding layer channel-wise. Then the P-CSP module performs feature integration.

\subsubsection{Neighborhood Voting Module}
At the output layer, the system generates two confidence values ($confs$), along with $label_{nei}$ and $label_{logits}$ for each data point. The algorithm first selects the $label_{logits}$ with the higher confidence. It then compares $label_{nei}$ with $label_{logits}$ to identify neighboring points with the same label. The final category is determined by aggregating the predicted label probabilities of these neighbors.

\subsection{MCAE Module}
The proposed MCAE module, illustrated in Fig.~\ref{fig:architecture}, has been specifically designed to fully exploit multidimensional feature information in a cascaded manner. In the initial stage, a centroid point is selected to obtain its coordinate information ${P_i}$ $(x, y, z)$ and color information ${F_i}$ $(R, G, B)$ . Subsequently, the KNN method is employed to ascertain the coordinate data for the corresponding $k$ neighboring points, represented as ${P_i}^k$, and the color information, represented as ${F_i}^k$. Next, the position and color information are merged along the channel dimension. The adjacent feature matrix of each point is encoded as follows:

\begin{flalign}\label{eq3}
	& \ {P_{i}^r} = FC(\{ {P_i}^k\}  \oplus ({P_i} - \{ {P_i}^k\} ) \oplus \left\| {{P_i} - \{ {P_i}^k\} } \right\|) &
\end{flalign}

\begin{flalign}\label{eq4}
	& \  {F_{i}^r} = FC(\{ {F_i}^k\}  \oplus ({F_i} - \{ {F_i}^k\} )) &
\end{flalign}

\noindent
where $P_{i}^r$ denotes multidimensional relative position information and $F_{i}^r$ denotes multidimensional relative color information. $FC$ represents delivering multidimensional information through a fully-connected layer. $ \oplus $ is the concatenation operation which aggregate all multidimensional information. $\left\|  \cdot  \right\|$ calculates the Euclidean distance between the neighboring points and the centroid point.

As illustrated in the Fig.~\ref{fig:architecture}, the objective is to enhance the network depth and expand the training space. We propose an enhanced point cloud-based residual network that is analogous to ResNet \cite{resnet}. Furthermore, the KNN and reshape methods are employed to incorporate local features.  Ultimately, max pooling is employed to downsample the features, ensuring their suitability for the subsequent scale layer. The aforementioned series of operations results in the formation of a cascading loop of original features at varying scales. 

For the entire feature ${Fea_i}$, shown in the MCAE module, an initial process is performed in a CBL layer to increase the depth of the model, followed by concatenation with ${P_i} $ and ${F_i} $ before the Max Pooling process. This allows the feature set to contain both coordinate and color local features. Subsequently, a straightforward attention learning process is conducted on the encoded ${Fea_i}$ feature, as outlined below:

\begin{flalign}\label{eq20}
	& \  S_{i}^{k}=softmax(FC(Fea_{i}^{k})) &
\end{flalign}
\begin{flalign}\label{eq21}
	& \  Fea_i=sum(Fea_{i}^{k}\bullet S_{i}^{k}) &
\end{flalign}

\noindent
where $sum$ is the point-wise summation of neighboring near points, and $\bullet$ is the matrix point multiplication. 

\subsection{P-CSP Module}
The decoding phase of the network primarily entails fusing sampled features at the same scale, maximizing the integration of valuable feature information across all scales. Motivated by the CSPNet \cite{csp} architecture, we introduce a P-CSP enhanced learning module designed to aggregate features across different scales, depicted in Fig.~\ref{fig:architecture}. After feature entry, the feature matrix is divided into two branches. One branch should utilize a CBL layer and propagate through residual backpropagation, while the other should pass through a CBL layer and learn directly via residual blocks. Notably, the number of feature channels for both branches is halved during the learning process. This module enhances gradient combinations while also preserving or enhancing the scale connectivity accuracy in the presence of residual backpropagation. Finally, KNN and reshape methods are utilized to incorporate local feature information into the features. The output results are obtained after passing through a CBL layer.

\section{Experiments}

\begin{figure}
	\begin{center}
		\includegraphics[width=0.5\textwidth]{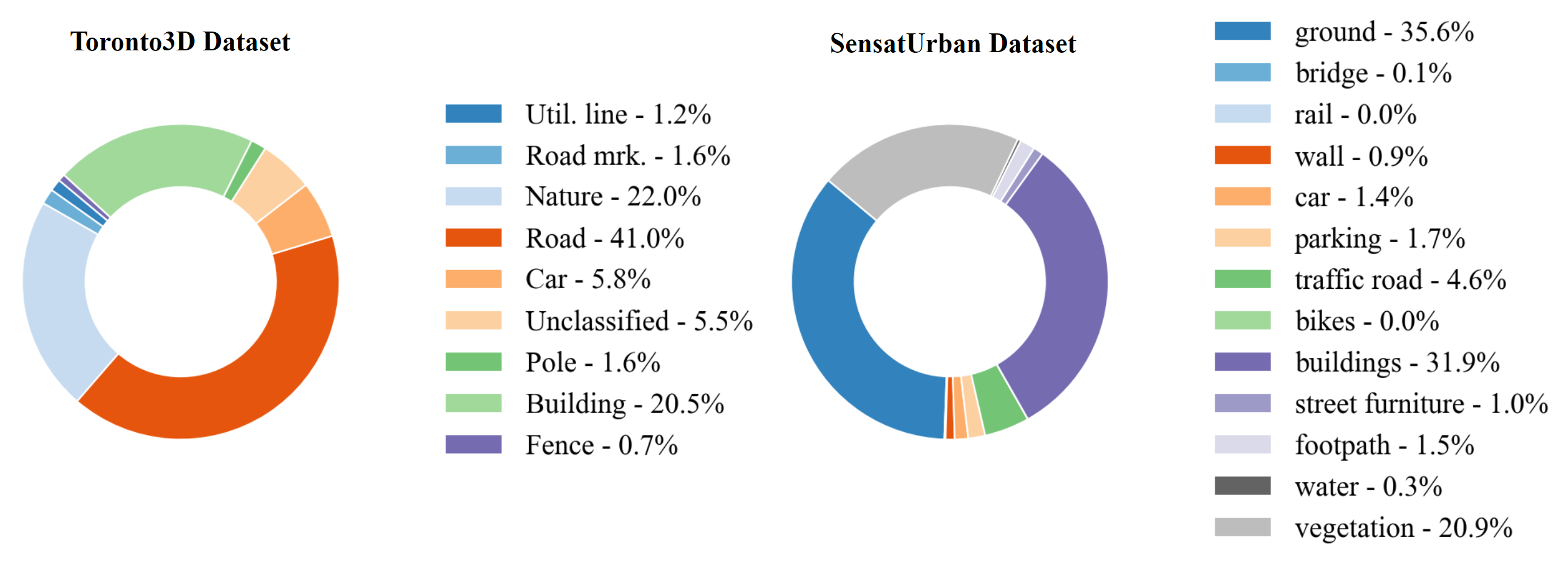}
	\end{center}
	\caption{Label distributions of Toronto3D and SensatUrban dataset.}
	\label{fig:label}
\end{figure}

We performed experiments on two public large-scale outdoor datasets, Toronto3D \cite{toronto} and SensatUrban \cite{sensaturban}, to evaluate the performance of the proposed architecture. The two datasets reflect the complex, real-world scenario of imbalanced object categories (Fig.~\ref{fig:label}). As for specific experiment settings, a total of 100 iterations were conducted using a NVIDIA RTX A6000 GPU, with the batch size of 4. To align with existing studies, we used overall accuracy (OA) and mean intersection over union (mIoU) to assess the performance of MCNet.

\begin{figure}
	\begin{center}
		\includegraphics[width=0.5\textwidth]{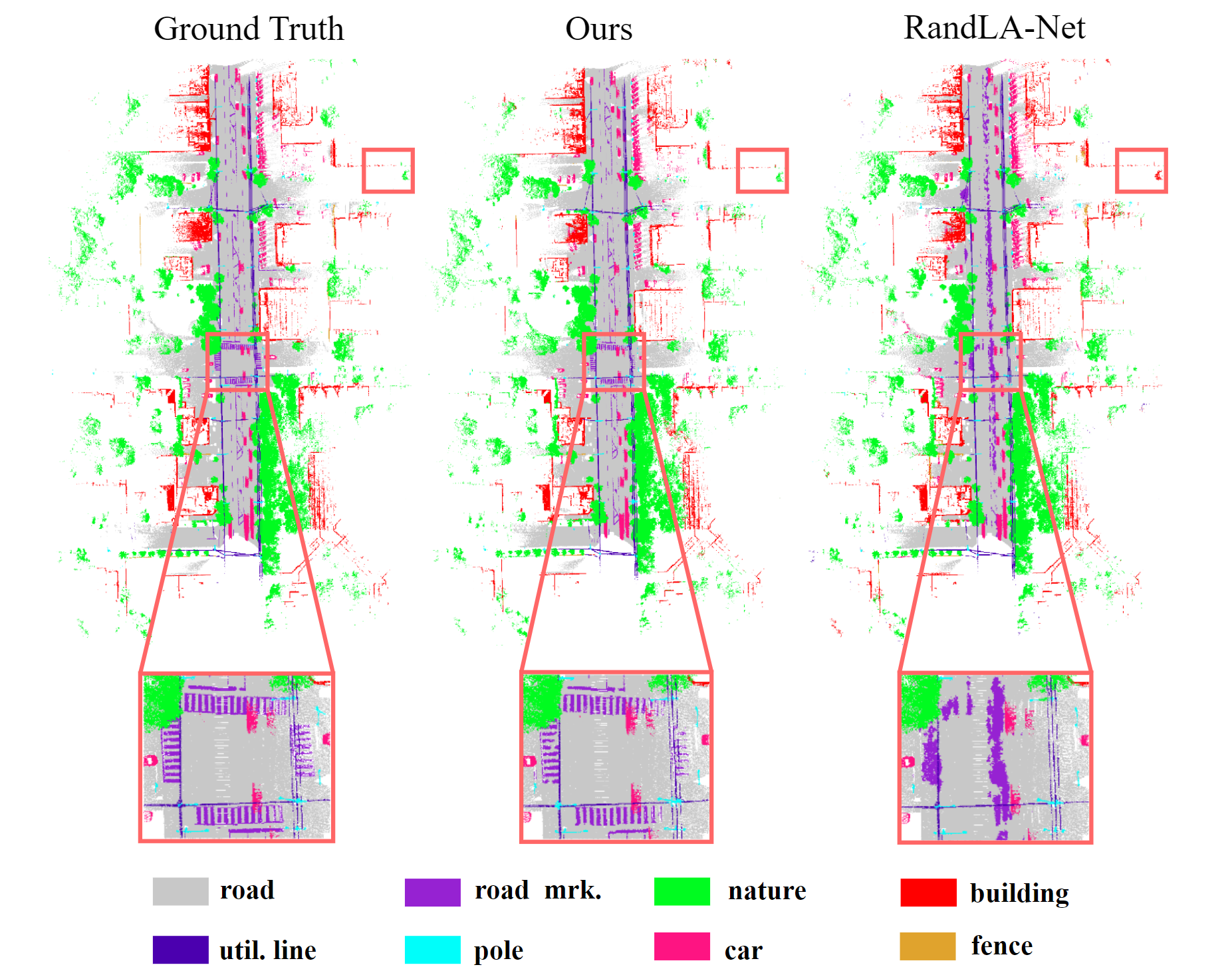}
	\end{center}
	\caption{Visual comparison of segmentation results between MCNet and the baseline RandLA-Net on the Toronto3D dataset.}
	\label{fig:five}
\end{figure}

\begin{table}[]
	\caption{The comparison of MCNet with other deep learning networks on the Toronto3D dataset using OA, and mIoU as evaluation metrics. The \textbf{bolded} values indicate the best performance for each metric}
	\resizebox{1.0\linewidth}{!}{
		\begin{tabular}{llllllllllll}
			\hline
			& Method                             & OA(\%)        & mIoU(\%)      & IoUs(\%)      &               &               &               &               &               &               &               \\ \cline{5-12}
			&                                    &               &               & road          & road m.       & nature       & build.        & util.l.       & pole          & car           & fence         \\ \hline
			& RandLA-Net \cite{randla}          & 94.3          & 81.7          & 96.6          & 64.2          & 96.9          & 94.2          & \textbf{88.0} & 77.8          & 93.3          & 42.8          \\
			& BAAF-Net \cite{baaf}              & 94.2          & 81.2          & 96.8          & \textbf{67.3}          & 96.8          & 92.2          & 86.8          & 82.3          & 93.1          & 34            \\
			& BAF-LAC \cite{baf}                & 95.2          & 82.2          & 96.6          & 64.7          & 96.4          & 92.8          & 86.1          & \textbf{83.9} & \textbf{93.7} & 43.5          \\
			& NeiEA-Net \cite{neiea}            & 97.0          & 80.9          & \textbf{97.1}          & 66.9          & 97.3          & 93.0          & 87.3          & 83.4          & 93.4          & 43.1          \\
			& TCFAP-Net  \cite{zhang2024tcfap}  & 97.0          & 81.9          & \textbf{97.1}          & 64.8          & 97.2          & \textbf{94.3} & 87.9          & 81.9          & 93.0          & 38.6          \\
			& LACV-Net \cite{zeng2024large}     & \textbf{97.4} & 82.7          & \textbf{97.1}          & 66.9          & 97.3          & 93.0          & 87.3          & 83.4          & 93.4          & 43.1          \\
			& \textbf{Ours}                      & \textbf{97.4} & \textbf{82.9} & 96.1          & 66.5          & \textbf{97.4} & 93.2          & 87.1          & 79.7          & 92.7          & \textbf{49.5} \\ \hline
		\end{tabular}
	}
	\label{Table:2}
\end{table}

\subsection{Experiment on the Toronto3D Dataset}

Table~\ref{Table:2} displays improvements of 3.1\% in OA and 1.2\% in mIoU compared with the baseline model RandLA-Net \cite{randla}. Moreover, MCNet outperforms other models in both the OA and mIoU metrics. In particular, MCNet exhibits the most effective segmentation capabilities for the categories ``nature" and ``fence". Fig.~\ref{fig:five} presents a comparative visualization of the validation results between MCNet and RandLA-Net .

Furthermore, MCNet demonstrates a noteworthy ability to segment objects in small samples. To illustrate, for categories such as ``fence", ``pole", and ``road m." which comprise less than 2\% of the total sample, MCNet attains improvements in IoU of 6.7\%, 1.9\%, and 1.3\%, respectively, in comparison to RandLA-Net. However, in the case of the ``util. l." category, RandLA-Net outperforms MCNet by 0.9\% in IoU and demonstrates the most optimal performance across all models.

\begin{table*}
	\caption{The comparison of MCNet with other deep learning networks on the SensatUrban dataset using OA, and mIoU as evaluation metrics. The \textbf{bolded} values indicate the best performance for each metric}
	\begin{center}
		\resizebox{1.0\linewidth}{!}{
			
			\begin{tabular}{@{}cccccccccccccccc@{}}
				\hline
				Method                              & OA($\% $)      & mIoU($\% $)    & IoUs($\% $)    &                &                &                &                &                &       &                &                &                &                &                &                \\ \cline{4-16}
				
				&                &                & ground         & veg            & building       & wall           & bridge         & parking        & rail  & traffic        & street         & car            & footpath       & bike           & water          \\
				\hline
				PointNet \cite{pointnet}           & 80.78          & 23.71          & 67.96          & 89.52          & 80.05          & 0.00           & 0.00           & 3.95           & 0.00  & 31.55          & 0.00           & 35.14          & 0.00           & 0.00           & 0.00           \\
				PointNet++ \cite{pointnet++}       & 84.30          & 32.92          & 72.46          & 94.24          & 84.77          & 2.72           & 2.09           & 25.79          & 0.00  & 31.54          & 11.42          & 38.84          & 7.12           & 0.00           & 56.93          \\
				SPGraph \cite{spg}                 & 76.97          & 37.29          & 69.93          & 94.55          & 88.87          & 32.83          & 12.58          & 15.77          & 15.48 & 30.63          & 22.96          & 56.42          & 0.54           & 0.00           & 44.24          \\
				SparseConv \cite{sparseconv}       & 85.27          & 42.66          & 74.10          & 97.90          & 94.20          & 63.30          & 7.50           & 24.20          & 0.00  & 30.10          & 34.00          & 74.40          & 0.00           & 0.00           & 54.80          \\
				KPConv \cite{kpconv}               & 93.20          & 57.58          & 87.10          & \textbf{98.91} & 95.33          & \textbf{74.40} & 28.69          & 41.38          & 0.00  & 55.99          & \textbf{54.43} & \textbf{85.67} & 40.39          & 0.00           & \textbf{86.30} \\
				BAAF-Net \cite{baaf}               & 91.70          & 59.60          & 80.00          & 93.90          & 94.00          & 65.70          & 25.40          & \textbf{63.00} & \textbf{50.20} & 61.90          & 42.40          & 79.70          & 42.10          & 0.00           & 79.60          \\
				SCF-Net \cite{scf}                 & 91.71          & 61.30          & 78.30          & 90.90          & 92.40          & 64.30          & 35.20          & 57.00          & 47.50 & \textbf{63.50} & 44.50          & 78.40          & 44.20          & 16.20          & 83.90          \\
				RandLA-Net \cite{randla}           & 89.78          & 52.69          & 80.11          & 98.07          & 91.58          & 48.88          & 40.75          & 51.62          & 0.00  & 56.67          & 33.23          & 80.14          & 32.63          & 0.00           & 71.31          \\
				NeiEA-Net \cite{neiea}             & 91.70          & 57.0           & 83.30          & 98.10          & 93.40          & 50.10          & \textbf{61.30} & 57.80          & 0.00  & 60.00          & 41.60          & 82.40          & 42.10          & 0.00           & 71.00          \\
				MVP-Net \cite{mvpnet}              & 93.30          & 59.40          & 85.10          & 98.50          & \textbf{95.90} & 66.60          & 57.50          & 52.70          & 0.00  & 61.90          & 49.70          & 81.80          & 43.90          & 0.00           & 78.20          \\
				PVCFormer-SA \cite{zhang2024point} & 93.80          & 62.40          & 83.50          & 97.50          & 94.40          & 68.30          & 51.60          & 51.70          & \textbf{50.20} & 58.40          & 48.60          & 79.80          & \textbf{47.60} & 0.00           & 79.40          \\
				DG-Net  \cite{liu2024semantic}     & 92.10          & 59.80          & 84.90          & 98.20          & 92.80          & 57.00          & 50.90          & 57.50          & 28.10 & 61.20          & 43.80          & 81.70          & 42.10          & 0.20           & 79.40          \\
				\textbf{Ours}                       & \textbf{94.00} & \textbf{64.50} & \textbf{87.60} & 98.30          & 95.10          & 60.00          & 59.20          & 57.90          & 48.90 & 58.30          & 44.90          & 81.60          & 43.20          & \textbf{24.80} & 81.40          \\
				\hline
			\end{tabular}
		}
	\end{center}
	
	\label{Table:3}
\end{table*}

\begin{figure}
	\begin{center}
		\includegraphics[width=0.5\textwidth]{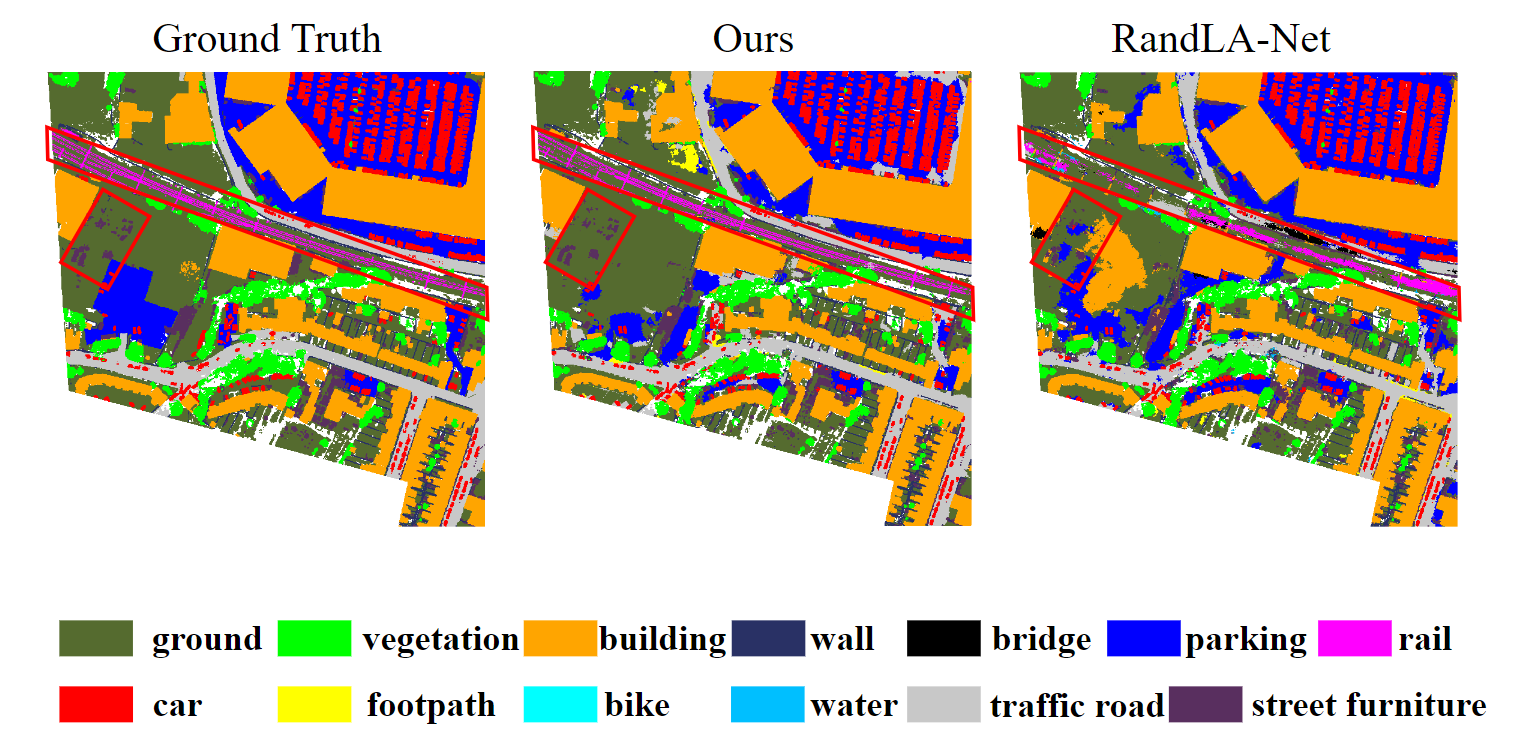}
	\end{center}
	\caption{Visual comparison of segmentation results between MCNet and the baseline RandLA-Net on the SensatUrban dataset.}
	\label{fig:four}
\end{figure}

\subsection{Experiment on the SensatUrban Dataset} 

As indicated in Table~\ref{Table:3}, our work on semantic segmentation has yielded impressive results, outperforming other models in OA ($94.0\% $) and mIoU ($64.5\% $) by a wide margin. Compared to the RandLA-Net, MCNet achieved significant improvements, with increases of 4.22\% in OA and 11.81\% in mIoU. Fig.~\ref{fig:four} displays the visualization results between MCNet and RandLA-Net. 

\begin{table}[!ht]
	\centering
	\caption{Segmentation performance comparison between MCNet and RandLA-Net on small-sample object categories in the SensatUrban dataset. The values represent the Intersection over Union (IoU) for each category, with the percentage symbol omitted for clarity}
	\resizebox{.99\columnwidth}{!}{
		\large 
		\begin{tabular}{llllllllll}
			\hline
			~ & bridge & rail & wall & car & parking & bike & street & footpath & water \\ 
			Proportion & 0.1\% & 0.0\% & 0.9\% & 1.4\% & 1.0\% & 0.0\% & 1.0\% & 1.5\% & 0.3\% \\ 
			\hline
			RandLA-Net \cite{randla} & 40.75 & 0 & 48.88 & 80.14 & 51.62 & 0 & 33.23 & 32.63 & 71.31 \\ 
			MCNet & 59.2 & 48.9 & 60 & 81.6 & 57.9 & 24.8 & 44.9 & 43.2 & 81.4 \\ 
			~ & $\uparrow$18.45 & $\uparrow$48.9 & $\uparrow$11.12 & $\uparrow$1.46 & $\uparrow$6.28 & $\uparrow$24.8 & $\uparrow$11.67 & $\uparrow$10.57 & $\uparrow$10.09 \\ \hline
		\end{tabular}
	}
	\label{tab:small-sample}
\end{table}

MCNet exhibits remarkable capabilities in the segmentation of small-sample objects. In the nine categories comprising less than 2\% of the total samples in SensatUrban, MCNet exhibits an average improvement of 15.9\% in segmentation IoU relative to RandLA-Net (see Table~\ref{tab:small-sample}). It is noteworthy that the majority of models encounter difficulties in segmenting the ``bike" category (see Table~\ref{Table:3}). In contrast, MCNet achieves a segmentation IoU of 24.80\%, a performance that is markedly superior to that of other models. This represents an improvement of 8.6\% over the second-best model, SCF-Net.

\begin{table}[!ht]
	\centering
	\caption{ablation study of mcnet. "SWS" means semantic-based weighted point sampling, "NV" denotes neighborhood voting}
	\begin{tabular}{c|cccc|c}
		\toprule
		Model & SWS & MCAE & P-CSP & NV & mIoU (\%) \\ 
		\midrule
		A & ~ & \checkmark & \checkmark & \checkmark & 61.18 \\ 
		B & \checkmark & ~ & \checkmark & \checkmark & 56.93 \\ 
		C & \checkmark & \checkmark & ~ & \checkmark & 63.48 \\ 
		D & \checkmark & \checkmark & \checkmark & ~ & 62.45 \\ 
		E & \checkmark & \checkmark & \checkmark & \checkmark & 64.50 \\ 
		\bottomrule
	\end{tabular}
	\label{tab:ablation}
\end{table}

\begin{table}[ht]
	\centering
	\caption{Ablation study of KNN with different number of neighbors}
	\label{tab:knn_ablation}
	\begin{tabular}{ccc}
		\toprule
		Number of Neighbors & OA (\%) & mIoU (\%) \\
		\midrule
		K = 9  & 93.92 & 64.43 \\
		K = 16 & 93.83 & 64.12 \\
		K = 25 & 94.00 & 64.50 \\
		K = 36 & 93.87 & 64.33 \\
		\bottomrule
	\end{tabular}
	
\end{table}

\subsection{Ablation Study}

For demonstrating the effectiveness of our proposed modules, we conduct ablation study on the SensatUrban \cite{sensaturban} dataset. The specific results are shown in the Table \ref{tab:ablation}

\subsubsection{Effectiveness of semantic-based weighted point sampling}The module for sampling based on semantic weights was removed in Model A and replaced with unweighted random sampling. The consequence of this operation is a reduction in the learning rate of small samples during the learning process, which in turn results in a decline in the performance of the model.

\subsubsection{Effectiveness of MCAE} We removed the proposed MCAE module in Model B and replaced it with the local feature aggregation (LAF) of RandLA-Net \cite{randla}. In theory, MCAE is capable of more effectively utilizing initial features and possesses a stronger local feature learning ability. As illustrated in Table \ref{tab:ablation}, the removal of the MCAE has the most pronounced impact, underscoring its unique capacity for point feature learning.

\subsubsection{Effectiveness of P-CSP} We replaced the P-CSP with MLP in Model C to test the contribution of the cross-stage feature aggregation module. Table~\ref{tab:ablation} indicates that replacing the P-CSP with MLP led to a slight drop in performance.

\subsubsection{Effectiveness of Neighborhood Voting} We removed the neighborhood voting module and made Model D directly output the semantic results of each point. This operation reduces the structural constraints on each point to a certain extent and does not allow the information of adjacent points to be combined in a more complete manner.

\subsubsection{Number of Neighbors in KNN}  We also investigated the impact of the value of K in the KNN algorithm within the MCAE module on the network's performance. As shown in Table \ref{tab:knn_ablation}, our experiments show that the network achieves optimal performance when K is set to be 25. Increasing K to 36 or decreasing K to 9 both lead to a slight decrease in performance.

\section{Conclusion}

This paper describes the design of a Multilateral Cascading Network (MCNet) for the processing of outdoor, large-scale point clouds. The network employs a semantic-weighted sampling technique to enhance the attention on small-sample objects. Subsequently, a multilateral cascading encoding module is devised to effectively capture the structural characteristics of neighboring regions and integrate semantic data. Moreover, a point cross-stage partial module is proposed for feature fusion. Ultimately, the output is augmented through neighborhood voting. Our research demonstrated that this approach yielded superior performance compared to other methods on two large-scale point cloud datasets, particularly excelling in semantic categories with small sample sizes.

\bibliographystyle{IEEEtran}
\bibliography{IEEEabrv,refs}

\begin{thebibliography}{10}
\providecommand{\url}[1]{#1}
\csname url@samestyle\endcsname
\providecommand{\newblock}{\relax}
\providecommand{\bibinfo}[2]{#2}
\providecommand{\BIBentrySTDinterwordspacing}{\spaceskip=0pt\relax}
\providecommand{\BIBentryALTinterwordstretchfactor}{4}
\providecommand{\BIBentryALTinterwordspacing}{\spaceskip=\fontdimen2\font plus
\BIBentryALTinterwordstretchfactor\fontdimen3\font minus
  \fontdimen4\font\relax}
\providecommand{\BIBforeignlanguage}[2]{{%
\expandafter\ifx\csname l@#1\endcsname\relax
\typeout{** WARNING: IEEEtran.bst: No hyphenation pattern has been}%
\typeout{** loaded for the language `#1'. Using the pattern for}%
\typeout{** the default language instead.}%
\else
\language=\csname l@#1\endcsname
\fi
#2}}
\providecommand{\BIBdecl}{\relax}
\BIBdecl

\bibitem{qi2021offboard}
C.~R. Qi, Y.~Zhou, M.~Najibi, P.~Sun, K.~Vo, B.~Deng, and D.~Anguelov,
  ``Offboard 3d object detection from point cloud sequences,'' in
  \emph{Proceedings of the IEEE/CVF Conference on Computer Vision and Pattern
  Recognition}, 2021, pp. 6134--6144.

\bibitem{moyano2021semantic}
J.~Moyano, J.~Le{\'o}n, J.~E. Nieto-Juli{\'a}n, and S.~Bruno, ``Semantic
  interpretation of architectural and archaeological geometries: Point cloud
  segmentation for hbim parameterisation,'' \emph{Automation in Construction},
  vol. 130, p. 103856, 2021.

\bibitem{el2021indoor}
N.~El-Sheimy and Y.~Li, ``Indoor navigation: State of the art and future
  trends,'' \emph{Satellite Navigation}, vol.~2, no.~1, p.~7, 2021.

\bibitem{pointnet}
C.~R. Qi, H.~Su, K.~Mo, and L.~J. Guibas, ``Pointnet: Deep learning on point
  sets for 3d classification and segmentation,'' in \emph{Proceedings of the
  IEEE conference on computer vision and pattern recognition}, 2017, pp.
  652--660.

\bibitem{pointnet++}
C.~R. Qi, L.~Yi, H.~Su, and L.~J. Guibas, ``Pointnet++: Deep hierarchical
  feature learning on point sets in a metric space,'' \emph{Advances in neural
  information processing systems}, vol.~30, 2017.

\bibitem{dgcnn}
Y.~Wang, Y.~Sun, Z.~Liu, S.~E. Sarma, M.~M. Bronstein, and J.~M. Solomon,
  ``Dynamic graph cnn for learning on point clouds,'' \emph{Acm Transactions On
  Graphics (tog)}, vol.~38, no.~5, pp. 1--12, 2019.

\bibitem{pointnext}
G.~Qian, Y.~Li, H.~Peng, J.~Mai, H.~Hammoud, M.~Elhoseiny, and B.~Ghanem,
  ``Pointnext: Revisiting pointnet++ with improved training and scaling
  strategies,'' \emph{Advances in Neural Information Processing Systems},
  vol.~35, pp. 23\,192--23\,204, 2022.

\bibitem{spg}
L.~Landrieu and M.~Simonovsky, ``Large-scale point cloud semantic segmentation
  with superpoint graphs,'' in \emph{Proceedings of the IEEE conference on
  computer vision and pattern recognition}, 2018, pp. 4558--4567.

\bibitem{randla}
Q.~Hu, B.~Yang, L.~Xie, S.~Rosa, Y.~Guo, Z.~Wang, N.~Trigoni, and A.~Markham,
  ``Randla-net: Efficient semantic segmentation of large-scale point clouds,''
  in \emph{Proceedings of the IEEE/CVF Conference on Computer Vision and
  Pattern Recognition}, 2020, pp. 11\,108--11\,117.

\bibitem{kpconv}
H.~Thomas, C.~R. Qi, J.-E. Deschaud, B.~Marcotegui, F.~Goulette, and L.~J.
  Guibas, ``Kpconv: Flexible and deformable convolution for point clouds,'' in
  \emph{Proceedings of the IEEE/CVF international conference on computer
  vision}, 2019, pp. 6411--6420.

\bibitem{neiea}
Y.~Xu, W.~Tang, Z.~Zeng, W.~Wu, J.~Wan, H.~Guo, and Z.~Xie, ``Neiea-net:
  Semantic segmentation of large-scale point cloud scene via neighbor
  enhancement and aggregation,'' \emph{International Journal of Applied Earth
  Observation and Geoinformation}, vol. 119, p. 103285, 2023.

\bibitem{leard}
Z.~Zeng, Y.~Xu, Z.~Xie, W.~Tang, J.~Wan, and W.~Wu, ``Leard-net: Semantic
  segmentation for large-scale point cloud scene,'' \emph{International Journal
  of Applied Earth Observation and Geoinformation}, vol. 112, p. 102953, 2022.

\bibitem{zeng2024large}
Z.~Zeng, Y.~Xu, Z.~Xie, W.~Tang, J.~Wan, and W.~Wu, ``Large-scale point cloud
  semantic segmentation via local perception and global descriptor vector,''
  \emph{Expert Systems with Applications}, vol. 246, p. 123269, 2024.

\bibitem{lai2022stratified}
X.~Lai, J.~Liu, L.~Jiang, L.~Wang, H.~Zhao, S.~Liu, X.~Qi, and J.~Jia,
  ``Stratified transformer for 3d point cloud segmentation,'' in
  \emph{Proceedings of the IEEE/CVF conference on computer vision and pattern
  recognition}, 2022, pp. 8500--8509.

\bibitem{10273676}
B.~Guo, L.~Deng, R.~Wang, W.~Guo, A.~H.-M. Ng, and W.~Bai, ``Mctnet: Multiscale
  cross-attention-based transformer network for semantic segmentation of
  large-scale point cloud,'' \emph{IEEE Transactions on Geoscience and Remote
  Sensing}, vol.~61, pp. 1--20, 2023.

\bibitem{resnet}
K.~He, X.~Zhang, S.~Ren, and J.~Sun, ``Deep residual learning for image
  recognition,'' in \emph{Proceedings of the IEEE conference on computer vision
  and pattern recognition}, 2016, pp. 770--778.

\bibitem{csp}
C.-Y. Wang, H.-Y.~M. Liao, Y.-H. Wu, P.-Y. Chen, J.-W. Hsieh, and I.-H. Yeh,
  ``Cspnet: A new backbone that can enhance learning capability of cnn,'' in
  \emph{Proceedings of the IEEE/CVF conference on computer vision and pattern
  recognition workshops}, 2020, pp. 390--391.

\bibitem{toronto}
W.~Tan, N.~Qin, L.~Ma, Y.~Li, J.~Du, G.~Cai, K.~Yang, and J.~Li, ``Toronto-3d:
  A large-scale mobile lidar dataset for semantic segmentation of urban
  roadways,'' in \emph{Proceedings of the IEEE/CVF conference on computer
  vision and pattern recognition workshops}, 2020, pp. 202--203.

\bibitem{sensaturban}
Q.~Hu, B.~Yang, S.~Khalid, W.~Xiao, N.~Trigoni, and A.~Markham, ``Towards
  semantic segmentation of urban-scale 3d point clouds: A dataset, benchmarks
  and challenges,'' in \emph{Proceedings of the IEEE/CVF conference on computer
  vision and pattern recognition}, 2021, pp. 4977--4987.

\bibitem{baaf}
S.~Qiu, S.~Anwar, and N.~Barnes, ``Semantic segmentation for real point cloud
  scenes via bilateral augmentation and adaptive fusion,'' in \emph{Proceedings
  of the IEEE/CVF Conference on Computer Vision and Pattern Recognition}, 2021,
  pp. 1757--1767.

\bibitem{baf}
H.~Shuai, X.~Xu, and Q.~Liu, ``Backward attentive fusing network with local
  aggregation classifier for 3d point cloud semantic segmentation,'' \emph{IEEE
  Transactions on Image Processing}, vol.~30, pp. 4973--4984, 2021.

\bibitem{zhang2024tcfap}
J.~Zhang, Z.~Jiang, Q.~Qiu, and Z.~Liu, ``Tcfap-net: Transformer-based
  cross-feature fusion and adaptive perception network for large-scale point
  cloud semantic segmentation,'' \emph{Pattern Recognition}, p. 110630, 2024.

\bibitem{sparseconv}
B.~Graham, M.~Engelcke, and L.~Van Der~Maaten, ``3d semantic segmentation with
  submanifold sparse convolutional networks,'' in \emph{Proceedings of the IEEE
  conference on computer vision and pattern recognition}, 2018, pp. 9224--9232.

\bibitem{scf}
S.~Fan, Q.~Dong, F.~Zhu, Y.~Lv, P.~Ye, and F.-Y. Wang, ``Scf-net: Learning
  spatial contextual features for large-scale point cloud segmentation,'' in
  \emph{Proceedings of the IEEE/CVF Conference on Computer Vision and Pattern
  Recognition}, 2021, pp. 14\,504--14\,513.

\bibitem{mvpnet}
H.~Li, H.~Guan, L.~Ma, X.~Lei, Y.~Yu, H.~Wang, M.~R. Delavar, and J.~Li,
  ``Mvpnet: A multi-scale voxel-point adaptive fusion network for point cloud
  semantic segmentation in urban scenes,'' \emph{International Journal of
  Applied Earth Observation and Geoinformation}, vol. 122, p. 103391, 2023.

\bibitem{zhang2024point}
S.~Zhang, B.~Wang, Y.~Chen, S.~Zhang, and W.~Zhang, ``Point and voxel cross
  perception with lightweight cosformer for large-scale point cloud semantic
  segmentation,'' \emph{International Journal of Applied Earth Observation and
  Geoinformation}, vol. 131, p. 103951, 2024.

\bibitem{liu2024semantic}
T.~Liu, T.~Ma, P.~Du, and D.~Li, ``Semantic segmentation of large-scale point
  cloud scenes via dual neighborhood feature and global spatial-aware,''
  \emph{International Journal of Applied Earth Observation and Geoinformation},
  vol. 129, p. 103862, 2024.

\end{thebibliography}
\end{document}